\documentclass[a4paper,fleqn]{cas-sc}
\usepackage{nomencl}
\makenomenclature
\usepackage{subfigure}
\usepackage{threeparttable}
\usepackage{graphicx}
\usepackage{float}
\usepackage{algorithm}
\usepackage{threeparttable}
\usepackage{hyperref}
\usepackage{url}
\usepackage{diagbox}
\usepackage{ulem}
\usepackage{caption}
\usepackage{xcolor}
\usepackage{lineno}
\usepackage{multicol}
\usepackage{etoolbox}
\usepackage{amsmath}
\usepackage{amssymb}
\usepackage{algpseudocode}
\renewcommand\nomgroup[1]{%
  \item[\bfseries
  \ifstrequal{#1}{A}{Generic variable}{%
  \ifstrequal{#1}{B}{\text{$\alpha$}-SPIDL}{%
  \ifstrequal{#1}{C}{\text{$\cal B$}-SPIDL}{}}}%
]}
\usepackage[authoryear]{natbib}

\def\tsc#1{\csdef{#1}{\textsc{\lowercase{#1}}\xspace}}
\tsc{WGM}
\tsc{QE}
\tsc{EP}
\tsc{PMS}
\tsc{BEC}
\tsc{DE}

\begin{document}
\begin{sloppypar}
\let\WriteBookmarks\relax
\def\floatpagepagefraction{1}
\def\textpagefraction{.001}
\shorttitle{Transportmetrica B: Transport Dynamics}
\shortauthors{Li, Wang and Zou et~al.}
\title [mode = title]{Physics-informed deep operator network for traffic state estimation} 

\author[1,2]{Zhihao Li}
\ead{2410791@tongji.edu.cn}

\author[1,2]{Ting Wang}[orcid=0000-0002-0460-0682]
\cormark[1]
\ead{2110763@tongji.edu.cn}

\author[1,2]{Guojian Zou}
\ead{2010768@tongji.edu.cn}

\author[1,2]{Ruofei Wang}
\ead{2410870@tongji.edu.cn}

\author[1,2]{Ye Li}
\ead{JamesLI@tongji.edu.cn}

\address[1]{The Key Laboratory of Road and Traffic Engineering, Ministry of Education, Tongji University, Shanghai 201804, China}
\address[2]{College of Transportation Engineering, Tongji University, Shanghai 201804, China}

\begin{abstract}
Traffic state estimation (TSE) fundamentally involves solving high-dimensional spatiotemporal partial differential equations (PDEs) governing traffic flow dynamics from limited, noisy measurements. While Physics-Informed Neural Networks (PINNs) enforce PDE constraints point-wise, this paper adopts a physics-informed deep operator network (PI-DeepONet) framework that reformulates TSE as an operator learning problem. Our approach trains a parameterized neural operator that maps sparse input data to the full spatiotemporal traffic state field, governed by the traffic flow conservation law. Crucially, unlike PINNs that enforce PDE constraints point-wise, PI-DeepONet integrates traffic flow conservation model and the fundamental diagram directly into the operator learning process, ensuring physical consistency while capturing congestion propagation, spatial correlations, and temporal evolution. Experiments on the NGSIM dataset demonstrate superior performance over state-of-the-art baselines. Further analysis reveals insights into optimal function generation strategies and branch network complexity. Additionally, the impact of input function generation methods and the number of functions on model performance is explored, highlighting the robustness and efficacy of proposed framework. 
\end{abstract}


\maketitle
\section{Introduction}
The rapid urbanization and increasing vehicular population have significantly intensified traffic-related challenges, including congestion, accidents, and environmental degradation \citep{zou2024mt,dantsuji2024hypercongestion}. Effective traffic management and control are therefore critical to mitigating these problems and ensuring sustainable transportation systems \citep{almukhalfi2024traffic}.
However, the foundation of efficient traffic management lies in the accurate and comprehensive understanding of real-time traffic states \citep{wang2022real}. Precise traffic state information is essential for enabling informed decision-making in traffic management and control \citep{wang2024koopman,korecki2024democratizing}. Unfortunately, obtaining such fine-grained traffic state data remains a significant challenge. On one hand, the deployment of traffic detection infrastructure, such as loop detectors, cameras, or radar, is often sparse due to high installation and maintenance costs \citep{xing2024urban}. This results in spatially limited data coverage. On the other hand, factors such as equipment failure, high maintenance requirements, adverse weather conditions, and other operational constraints can result in missing or noisy data. These limitations have prompted the development of traffic state estimation (TSE) technology, which aims to automatically fill the gaps in the spatiotemporal collection of traffic information on some roads \citep{seo2017traffic}. 

Accurate estimation of traffic states across a road network relies heavily on understanding the complex interplay of spatial and temporal dynamics. Traditional TSE approaches, such as physics-based models \citep{wang2005real,ngoduy2011low, nantes2016real} and statistical-based methods \citep{kyriacou2022bayesian,afrin2021probabilistic}, often struggle with the high-dimensional, nonlinear, and heterogeneous nature of traffic data, particularly when observations are sparse or irregularly sampled. In recent years, deep learning models have emerged as powerful tools for TSE, leveraging their ability to capture intricate spatiotemporal patterns without requiring explicit assumptions about underlying physical processes \citep{xu2020ge,boquet2020variational}. Deep learning methods are highly expressive and have few assumptions, but they are limited by high data requirements, computational complexity, and low interpretability. To bridge the gap between the rigorous physics of traditional models and the flexible pattern recognition capabilities of deep learning, a novel paradigm known as physics-informed neural networks (PINNs) has gained significant traction in TSE and related scientific computing fields \citep{wang2024expert,di2023physics}. Instead of relying solely on vast amounts of observational data to learn spatiotemporal mapping, PINNs constrain the solution of the neural network to be consistent with the underlying governing physics \citep{huang2024incorporating}. However, challenges remain, such as the choice of the most appropriate physics model, the computational cost of solving the coupled optimization problem, and ensuring robust training convergence. 

Fundamentally, the task of TSE boils down to solving complex, often high-dimensional, spatiotemporal Partial Differential Equations (PDEs) that govern traffic flow dynamics, given limited and noisy measurements. Contemporary deep learning approaches tackling this PDE-solving problem primarily fall into two major categories: PINNs and Neural Operators (NOs) \citep{thodi2024fourier}. While PINNs, as discussed, integrate physical laws as soft constraints during training, our focus now shifts to the Neural Operator paradigm. NOs have attracted attention due to their efficiency in learning complex mappings between function spaces \citep{kovachki2023neural}. Operator learning frameworks, such as the Deep Operator Network (DeepONet), have shown remarkable potential in modeling complex systems by learning functional mappings from input spaces to output spaces \citep{lu2021learning,wang2021learning}. Unlike conventional deep learning models that predict discrete outputs (e.g., scalar or vector values), DeepONet is designed to learn operators—mappings between infinite-dimensional function spaces. 

Therefore, this paper proposes a physics-informed deep operator network framework (PI-DeepONet) for TSE. This approach trains a parameterized operator that maps arbitrary input data to the solution of the underlying traffic flow conservation law. Crucially, this approach synergizes data-driven learning with physical constraints, compared to PINNs that rely on point-wise PDE residual constraints during training. Within our framework, traffic states are conceptualized as a spatiotemporal field governed by intrinsic dynamics, including congestion propagation, spatial correlations, and temporal evolution. The operator we aim to learn encapsulates these dynamics by directly modeling the functional mapping from spatiotemporal inputs to the complete state field. PI-DeepONet thus achieves dual objectives: (1) Learning the governing operator mapping from inputs to the complete state field; (2) Ensuring physical consistency by satisfying the flow conservation law. This provides a direct, physics-regularized mapping from raw observational data to continuous PDE solutions across road networks. The contributions of this paper are thus outlined as follows: 
\begin{itemize}
\item We reframe traffic state estimation as a continuous function mapping problem and introduce deep neural operators (DeepONet) to learn the mapping from sparse inputs to the full spatiotemporal state field. 
\item We integrate the traffic fundamental diagram and the traffic flow conservation equation into the DeepONet architecture and construct a TSE framework based on PI-DeepONet. 
\item We conduct experiments on NGSIM dataset, demonstrating the superiority of PI-DeepONet over baselines. Further, We explored the impact of input function generation methods and the number of functions in the Branch network on model performance.
\end{itemize} 

The remainder of the paper is organized as follows. Section \ref{s2} reviews previous work on TSE and the application of neural operators in the field of transportation. Section \ref{s3} introduces the TSE framework based on DeepONet and PI-DeepONet in detail. The experimental settings and results are presented in Section \ref{s4}. Section \ref{s5} concludes our work and prospects. 

\section{Related work} \label{s2}
\subsection{Traffic state estimation methods}
Existing TSE methods are mainly categorized into model-driven, data-driven and hybrid-driven. Model-driven approaches to traffic flow analysis utilize physical models to achieve state estimation through the integration of real-time data \citep{wang2005real,ahmed2014significance}. Nonetheless, these methodologies are constrained by the inherent assumptions of the physical models employed, which can hinder their ability to accurately capture the intricate dynamic characteristics of actual traffic systems. In contrast, data-driven approaches leverage machine learning techniques to extract patterns and correlations from historical data without reliance on predefined traffic flow models \citep{xu2020ge,xu2021traffic,zhang2024data}. For instance, \citep{abdelraouf2022sequence} proposes a Seq2seq GCN-LSTM deep learning model to accurately estimate network-wide traffic volume and speed using sparse probe vehicle data. Their efficacy is heavily contingent upon the quality of the data and they often lack a comprehensive representation of the underlying physical mechanisms governing traffic flow, resulting in limited interpretability and robustness of the outcomes. To address this issue, researchers introduced PINN framework \citep{raissi2019physics}, which generates traffic estimates constrained by physical laws by incorporating a discretized macroscopic traffic flow model into the network architecture \citep{shi2023physics,huang2024incorporating}. For example, \citep{shi2021physics} designed a fundamental diagram learner and integrate it into the PINN to realize the joint optimization of state estimation, parameter calibration, and fundamental diagram estimation. \citep{zhang2024physics} have embedded base map parameters within a computational graph framework, combining this with PINN to enable the state reconstruction of entire roadways with limited observational data. \citep{wang2024knowledge} introduces  stochastic fundamental diagram  model into the PIDL architecture to effectively capture the scatter effect characteristics of traffic flow. \citep{xu2024traffic} proposed a novel MS-CIG model fuses fixed and mobile detector data using a cross-layer random walk and weighted spatiotemporal graph to accurately infer traffic state data for road sections without detectors. \citep{xue2024sparse}proposed a novel sparse mobile crowdsensing framework to enhance traffic state estimation accuracy by leveraging spatial and temporal correlations from limited vehicular data. \citep{wu2024traffic} proposed a novel Gaussian process-based method with a kernel rotation re-parametrization scheme to impute traffic state data. Nevertheless, the aforementioned methods remain fundamentally rooted in the conventional framework of network learning functions.

\subsection{The application of neural operators in the field of transportation}
NOs present innovative solutions for TSE learning the solution operators of PDEs pertinent to traffic flow. For example, \citep{gao2024comparative} conducted a comparative analysis of various NOs, including DeepONet and Fourier neural operator (FNO), utilizing both simulated and real traffic datasets, such as the NGSIM data. Their findings indicate that these models are adept at real-time traffic prediction, as they can adapt to varying initial and boundary conditions without necessitating repeated training. In scenarios characterized by sparse detection data, \citep{harting2025closed} developed a closed-loop FNO observer that integrates real-time sensor feedback, significantly improving the robustness of density estimation. Furthermore, the ON-Traffic framework introduced by \citep{rap2025traffic} uses Lagrangian mobile sensor data for online traffic flow estimation while incorporating uncertainty quantification, thus illustrating the potential of NOs in handling unstructured data environments. To enhance the physical consistency of the model, \citep{thodi2024fourier} proposed the Physical- Informed Fourier neural operator (PI-FNO), which improves the prediction abilities of the surge through regularization based on the Lighthill-Whitham-Richards (LWR) conservation law, and has been successfully applied to model density dynamics on urban signalized roadways. Beyond TSE, NOs have also demonstrated significant advancements in traffic control applications. For example, \citep{zhang2025mitigating} designed a boundary controller based on NOs for the Aw-Rascle-Zhang (ARZ) traffic model, achieving a 300-fold increase in the solution speed of the traditional backstepping control kernel while ensuring closed-loop stability through Lyapunov analysis. Furthermore, \citep{liu2024scalable} and \citep{chen2024physics} incorporated the PI-DeepONet into the resolution of mean-field game (MFG) equilibria, facilitating efficient generalization of autopilot speed control by training operators that are independent of initial conditions. Despite the promising results highlighted in these studies regarding the applicability of NOs to TSE challenges, there remains a notable gap in the literature, as previous research has rarely conceptualized TSE problems as mappings between continuous spatiotemporal fields. This oversight may restrict the functional mapping capabilities of NOs.

\section{Methodology} \label{s3}
\subsection{Problem definition}
TSE refers to the process of restoring the traffic state data of the entire spatiotemporal field of a roadway section through part of the spatiotemporal observation data of the roadway section under study. Specifically, the purpose of the study is the variables of traffic state $s$, including flow $q\left( {x,t} \right)$, speed $v\left( {x,t} \right)$, and density $\rho \left( {x,t} \right)$, for a section of the highway long $L$, where $x$ and $t$ represent the location and time of data collection, respectively. The complete spatiotemporal field is defined as $\Omega$, and the finite set of observed data is defined as a subset ${s_C} \in {s_\Omega }$. In the course of the later study, we choose speed $v\left( {x,t} \right)$ to use it as a traffic state for the study.

Previous TSEs have aimed to develop a spatiotemporal mapping function ${s_\Omega } = {f_\theta }\left( {{s_C}} \right)$ to facilitate reductions based on the observed data. In contrast, this paper seeks to learn operators $G:u\left( {x,t} \right) \to s\left( {x,t} \right)$ through the application of NOs, where $u\left( {x,t} \right)$ arbitrary spatiotemporal correlation functions are represented. It is important to note that these operators function between different functions, thereby elevating our problem from a point-to-point mapping framework to a function-to-function mapping paradigm.

\subsection{DeepONet for TSE}
\subsubsection{Overall architecture}
\begin{figure}[pos=htbp]
    \centering
\includegraphics[width=1\textwidth]{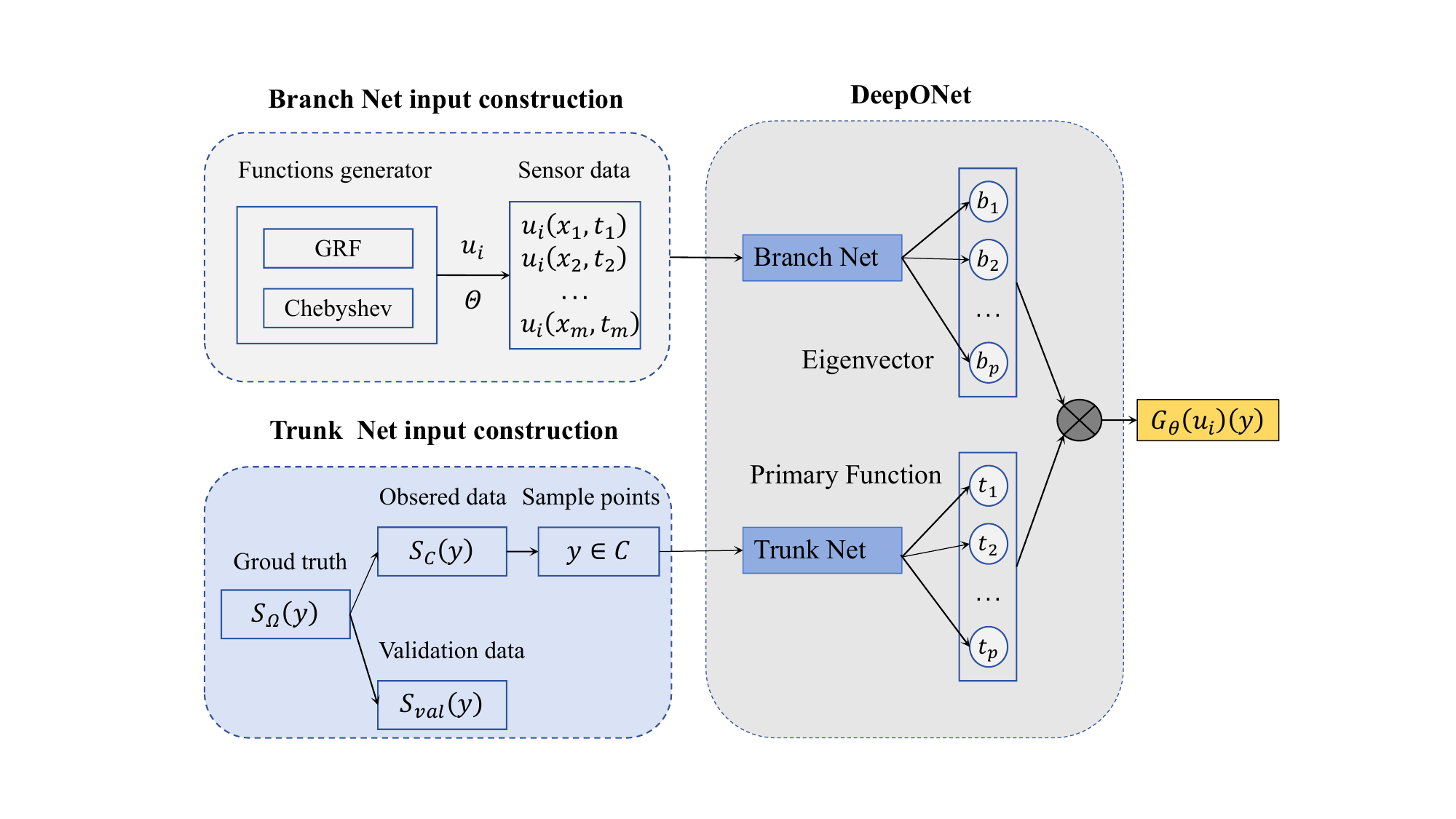}
    \caption{\centering{The Framework of DeepONet}}
	\label{DeepONet}
\end{figure}
The DeepONet architecture is shown in Fig. 1. DeepONet is designed to learn operators that map between distinct functions, as opposed to learning functions that relate different variables, such as integration, derivation, and partial differential equations. Specifically, with $u\left( {x,t} \right)$ and $s\left( {x,t} \right)$ denoting the input function and the objective function, respectively, the target of DeepONet is to learn the operators:
\begin{equation}
G:u\left( {x,t} \right) \to s\left( {x,t} \right) 
\end{equation}
In TSE, $u\left( {x,t} \right)$ is an arbitrary spacetime function, and $s\left( {x,t} \right)$ is a function of the transportation state variable with respect to spacetime. The theoretical basis of DeepONet is the universal approximation theorem for operators \citep{chen1995universal}, which states that, for arbitrary $\varepsilon  > 0$, there exist positive integers$n,p,m$, constants$c_i^k,W_{{b_{ij}}}^k,b_{{b_{ij}}}^k,{W_{tk}},{b_{tk}}$, such that:
\begin{equation}
    \left| {G\left( u \right)\left( y \right) - \sum\limits_{k = 1}^p {\sum\limits_{i = 1}^n {c_i^k\sigma \left( {\sum\limits_{j = 1}^m {W_{{b_{ij}}}^ku\left( {{x_j}} \right) + b_{{b_i}}^k} } \right) \cdot \sigma \left( {{W_{tk}} \cdot y + {b_{tk}}} \right)} } } \right| < \varepsilon 
\end{equation}
This theorem not only proves the feasibility of operator learning, but also shows the direction for NN-based operator learning. Inspired by the universal approximation theorem for operators, DeepONet model introduces two vertically stacked NNs to characterize the operator approximation theorem from the network level, which are called Branch network and Trunk network. The correspondence between Branch network and Trunk network with the universal approximation theorem is respectively as follows:
\begin{equation}
    Branch~net:\sum\limits_{i = 1}^n {c_i^k\sigma \left( {\sum\limits_{j = 1}^m {W_{{b_{ij}}}^ku\left( {{x_j}} \right) + b_{{b_i}}^k} } \right)} 
\end{equation}
\begin{equation}
    Trunk~net:\sum\limits_{k = 1}^p {\sigma \left( {{W_{tk}} \cdot y + {b_{tk}}} \right)}
\end{equation}
where $\sigma$ denotes the activation function, and $W_{{b_{ij}}}^k,{W_{tk}}$ and ${b_{tk}},b_{{b_{ij}}}^k$ denote the weights and biases of the Branch and Trunk networks, respectively.$x_j$ is the sampling region of the input function $u(x_j)$, which is called the configuration point. $y$ is the independent variable of the output function $s(y)$. It is important to acknowledge that this configuration implies that the input function may not have any correlation with the output function across any domain level. This premise underlies the rationale for employing arbitrary space-time functions as input functions in this study. Returning to the TSE problem, operator learning can be characterized as follows:
\begin{equation}
    {G_\theta }\left( u \right)\left( y \right) = \underbrace {g\left( {u\left( {{x_1},{t_1}} \right),u\left( {{x_2},{t_2}} \right),...,u\left( {{x_m},{t_m}} \right);{\theta _b}} \right)}_{Branch} \cdot \underbrace {f\left( {\left( y \right);{\theta _t}} \right)}_{Trunk}
\end{equation}
where ${G_\theta }$ are the operators actually represented by the networks, $g$ and $f$ representing Branch network and Trunk network, respectively. ${\theta _b}$,${\theta _t}$ representing the weight parameters and biases of the two networks, respectively. In the DeepONet model, the input of the Branch network is ${U_i} = {\left[ {{u_i}\left( {{x_1},{t_1}} \right),{u_i}\left( {{x_2},{t_2}} \right),...,{u_i}\left( {{x_m},{t_m}} \right)} \right]^{\rm T}}$, which is the value of the input function $u\left( {x,t} \right)$ at the configuration point ${\Theta } = \left[ {\left( {{x_1},{t_1}} \right),\left( {{x_2},{t_2}} \right),...,\left( {{x_m},{t_m}} \right)} \right]$, and the output is a feature vector $B=[b_1,b_2...,b_p]^T$ with the dimension of $p$. The input of the Trunk network is the independent variable $y$ of the spatiotemporal function of the traffic state in the target region, and the output is a set of basis function vectors $T=[t_1,t_2...,t_p]$ with the same dimension. The output of the two can be obtained by the dot product of the predicted value ${G_\theta }\left( u \right)\left( {y} \right)$ of the target function $s(y)$ in the region to be predicted, combined with the labeled value $s_C$ of supervised learning to achieve the purpose of the network learning operator.

\subsubsection{Training Sample Construction}

In the aforementioned expression, the input function $u\left( {x,t} \right)$ could be constructed using a random function field. This study examined both Gaussian Random Fields (GRF) (\cite{de1997bayesian}) and Chebyshev polynomials (\cite{mason2002chebyshev}), with their respective formulations and underlying principles detailed as follows, respectively:

\textbf{GRF:} A stochastic function whose core mathematical basis is an infinite dimensional extension of the Gaussian distribution, fully embedded through the mean and covariance functions, with the following core expression:
\begin{equation}
u\left( x \right) \sim \Gamma \left( {m\left( x \right),k\left( {{x_1},{x_2}} \right)} \right)
\end{equation}
where $\Gamma$  refers to the Gaussian process; $m(x)$ represents the expectation function of the random field, which is fixed to 0 in this study to ensure that there is no overall bias; and $k(x_1,x_2)$ represents the covariance function, which has the following expression:
\begin{equation}
    {k_l}\left( {{x_1},{x_2}} \right) = \exp \left( { - \frac{{{{\left| {{x_1} - {x_2}} \right|}^2}}}{{2{l^2}}}} \right)
\end{equation}
where $(x_1,x_2)$ is the randomly selected neighboring points, which are only used to determine the specific generating function. $l$ is the length scale, which is mainly used to control the smoothness of the function. When $l$ is larger, the covariance decays slower, the correlation of neighboring points is stronger, and the generated $u(x)$ is smoother; when $l$ is smaller, the covariance decays faster, the correlation of neighboring points is weaker, and $u(x)$ is more oscillating. In this study, the reference value of $l$ is taken to be 0.2, $(x_1,x_2)$ is randomly generated from the white noise field, and $\Gamma$ is realized by combining Gaussian filtering containing both spatial and temporal dimensions, while normalization is carried out to get the input function $u(x,t)$. The specific pseudo-code for the implementation is shown in Appendix algorithm \ref{alg:Generation of GRF}.

\textbf{Chebyshev polynomials:} A type of orthogonal polynomials whose mathematical basis is the theory of orthogonality and approximation of function spaces. Specifically, orthogonal functions possess the following properties:
\begin{equation}
    \int_a^b {{P_i}(x){P_j}(x)w(x)dx}    = \left\{ {\begin{array}{*{20}{l}}
{0,}&{i \ne j}\\
{{c_i} > 0,}&{i = j}
\end{array}} \right.
\end{equation}
where $c_i$ is the normalization constant and $w(x)$ is the weight function. $w(x)$ takes the values of $\frac{1}{{\sqrt {1 - {x^2}} }}$ and $\sqrt {1 - {x^2}}$, which correspond to the Chebyshev polynomials of the first and second classes, respectively. In this study, the Chebyshev polynomials of the first class are selected as the basis, the expression is as follows:
\begin{equation}
\int_{-1}^1 T_i(x)T_j(x) \cdot \frac{dx}{\sqrt{1 - x^2}} = 
\begin{cases}
0, & i \ne j \\
\frac{\pi}{2}, & i = j \ge 1 \\
\pi, & i = j = 0
\end{cases}
\end{equation}

\begin{equation}
    {T_0}(x) = 1,\quad {T_1}(x) = x,\quad {T_{n + 1}}(x) = 2x{T_n}(x) - {T_{n - 1}}(x)
\end{equation}
Based on the above, denoting by $T_i$ a set of orthogonal polynomials defined over the interval $[-1,1]$, the input function u is obtained by the following expression:
\begin{equation}
    u\left( x \right) = \sum\limits_{i = 0}^N {{a_i}{T_i}\left( {\frac{x}{M}} \right),x \in \left[ { - M,M} \right]} 
\end{equation}
where the coefficients $a_i$ are used for function generation and function complexity control , and the $M$ are used to map the interval $[-1,1]$ to$[-M,M]$. A series of functions $u(x,t)$ can be generated by random sampling $a_i$.The specific pseudo-code for the implementation is shown in Appendix algorithm \ref{alg:Generation of Chebyshev}
Note that this study uses random spatiotemporal functions as input functions for two main reasons. On the one hand, the use of randomly generated spatiotemporal functions can prove that input functions do not impose additional data requirements in TSE tasks based on the DeepONet framework. On the other hand, training with arbitrary spatiotemporal functions as input functions can theoretically produce models with good input function generalization.

After generating the function $u$, the values of the function at the configuration points $\Theta $ will be used as inputs to the Branch network. Configuration points $\Theta $ are a series of randomly selected spatiotemporal points used only to determine the inputs to the branching network and do not require any information. The specifics of the independent variables $y$ in the target region are determined by the research question. The labeled values $s(y)$ are usually already present in the dataset or can be obtained by numerical solving. Taken together, the samples required for DeepONet model training can be expressed as the following form:
\begin{equation}
    \left[ {u,y,s\left( y \right)} \right] = \left[ {\begin{array}{*{20}{c}}
{{u_1}\left( {{y_1}} \right),{u_1}\left( {{y_2}} \right),...,{u_1}\left( {{y_m}} \right)}\\
{{u_2}\left( {{y_1}} \right),{u_2}\left( {{y_2}} \right),...,{u_2}\left( {{y_m}} \right)}\\
 \vdots \\
{{u_n}\left( {{y_1}} \right),{u_n}\left( {{y_2}} \right),...,{u_n}\left( {{y_m}} \right)}
\end{array}} \right],\left[ {\begin{array}{*{20}{c}}
{{y_1}}\\
{{y_2}}\\
 \vdots \\
{{y_C}}
\end{array}} \right],\left[ {\begin{array}{*{20}{c}}
{s\left( {{y_1}} \right)}\\
{s\left( {{y_2}} \right)}\\
 \vdots \\
{s\left( {{y_C}} \right)}
\end{array}} \right]
\end{equation}
In TSE, the independent variables $y$ are essentially the spatiotemporal coordinates of the region to be predicted, and the labeled values $s(y)$ are the observed data $s_C$.

\subsubsection{Loss function of DeepONet}
Based on the above, the operator loss function of DeepONet can be given in the following form:
\begin{equation}
    {L_{operator}}\left( \theta  \right) = \frac{1}{{{N}{P}}}{\sum\limits_{i = 1}^{{N}} {\sum\limits_{j = 1}^{{P}} {\left| {{G_\theta }\left( {{u_i}} \right)\left( {{y_{ij}}} \right) - s\left( {{y_{ij}}} \right)} \right|} } ^2}
\end{equation}
where ${N}$ denotes the number of input functions and ${P}$ denotes the number of labeled points used for training. ${G_\theta }\left( u \right)\left( y \right)$ and $s\left( {{y_{ij}}} \right)$ denote the predicted and true values, respectively. Note that for a DeepONet model, the inputs needed for computation are only the value ${U_i} = {\left[ {{u_i}\left( {{x_1},{t_1}} \right),{u_i}\left( {{x_2},{t_2}} \right),...,{u_i}\left( {{x_m},{t_m}} \right)} \right]^{\rm T}}$ of an input function at $\Theta $ and the spatiotemporal region to be predicted, and the form of $N$ input functions is taken here mainly to describe the universal training process.

\subsection{PI-DeepONet for TSE}
\subsubsection{Overall architecture}
\begin{figure}[pos=htbp]
    \centering
\includegraphics[width=0.9\textwidth]{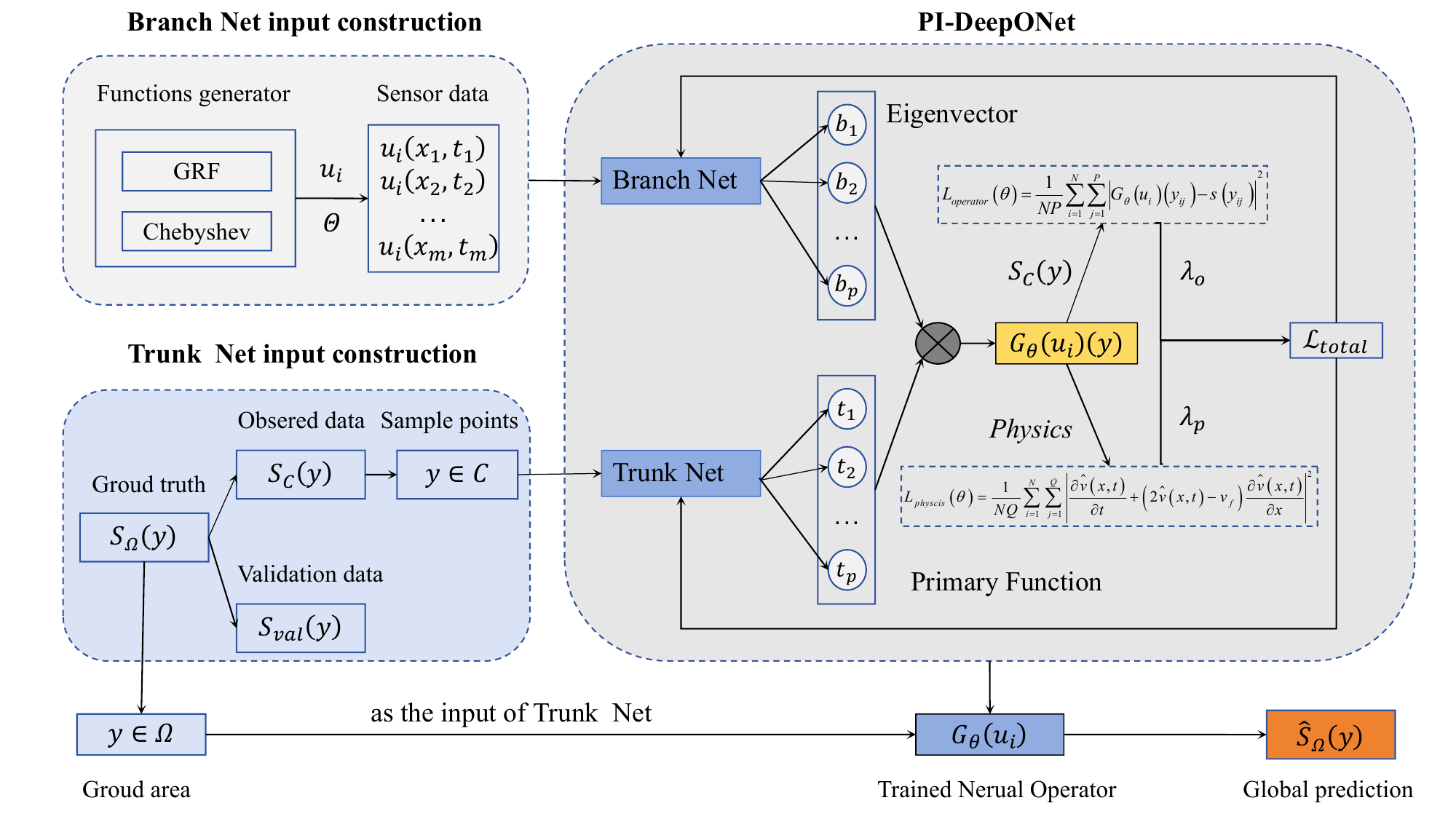}
    \caption{\centering{The Framework of PI-DeepONet}}
	\label{pi_DeepONet}
\end{figure}

The aforementioned DeepONet model effectively transforms the point-to-point mapping challenge inherent in the TSE problem into a function-to-function framework. However, the physical interpretation of the operator $G$, as learned by the network, remains ambiguous, which significantly impacts the model's interpretability. To address this limitation, we propose a physically-informed approach termed PI-DeepONet. The overarching framework of PI-DeepONet is illustrated in Fig.\ref{pi_DeepONet}, wherein a key enhancement involves the incorporation of the LWR-Greenshields model as a physical regularization term. This framework initially calibrates the parameters of the physical model using the training data, subsequently leading to the formulation of a physical loss function. The physical loss function is then integrated with the data loss function through weighting parameters ${\lambda _o},{\lambda _p}$ to establish the final loss function $L_{total}(\theta )$.

\subsubsection{Knowledge used for regularization training}
This section focuses on the process of introducing a physical model of traffic flow as a priori knowledge into an operator learning framework. The representative LWR model \citep{lighthill1955kinematic} and Greenshields fundamental diagram \citep{greenshields1935study} is considered in the proposed framework. Specifically, the LWR model gives a conservation law for traffic flow on a continuous unidirectional road:
\begin{equation}
\begin{array}{l}
{\partial _t}\rho  + {\partial _x}\left( {\rho v} \right) = 0\\
{\partial _t}\rho  + v{\partial _x}\rho  + \rho {\partial _x}v = 0
\end{array}
\end{equation}
combining the Greenshields model expression $v = {v_f}\left( {1 - {\rho  \mathord{\left/
 {\vphantom {\rho  {{\rho _m}}}} \right.
 \kern-\nulldelimiterspace} {{\rho _m}}}} \right)$ for the density-speed relationship gives the following expression:
\begin{equation}
 \frac{{{\rho _m}}}{{{v_f}}}\frac{{\partial v}}{{\partial t}} + v \cdot \frac{{{\rho _m}}}{{{v_f}}}\frac{{\partial v}}{{\partial x}} + \frac{{{\rho _m}\left( {v - {v_f}} \right)}}{{{v_f}}}\frac{{\partial v}}{{\partial x}} = 0
\end{equation}
\begin{equation}
\frac{{\partial v}}{{\partial t}} + \left( {2v - {v_f}} \right)\frac{{\partial v}}{{\partial x}} = 0
\end{equation}
where $v_f$ denotes the free stream speed of the roadway, which may be calibrated prior to the initiation of model training. The anticipated value of the velocity at this specific location can be articulated as $\widehat v\left( {x,t} \right) = {G_\theta }\left( {{u_i}} \right)\left( {x,t} \right)$. Given that the term on the right side of the equation is equal to 0, the residual associated with the prediction in accordance with the mass conservation principle can be formulated as follows:
\begin{equation}
    R_\theta ^i\left( {x,t} \right) = \frac{{\partial \widehat v\left( {x,t} \right)}}{{\partial t}} + \left( {2\widehat v\left( {x,t} \right) - {v_f}} \right)\frac{{\partial \widehat v\left( {x,t} \right)}}{{\partial x}}
\end{equation}
where $R_\theta ^i\left( {x,t} \right)$ denotes the residual of the selected $u_i$ input function at $(x,t)$ for the mass conservation law. This residual quantifies the extent to which the model's predictions deviate from the established mass conservation law. Consequently, a physical loss function can be formulated based on this deviation as follows:
\begin{equation}
    {L_{physcis}}\left( \theta  \right) = \frac{1}{{NQ}}\sum\limits_{i = 1}^N {\sum\limits_{j = 1}^Q {R_\theta ^i{{\left( {x,t} \right)}^2}} } 
\end{equation}
where $Q$ denotes the number of points at which the physical residuals are calculated.

\subsubsection{Combined loss function}
By combining the initial data loss of the DeepONet model, the total loss function can be formulated as follows:
\begin{equation}
\begin{array}{l}
{L_{total}}\left( \theta  \right)
 = {\lambda _o}{L_{operator}}\left( \theta  \right) + {\lambda _p}{L_{physcis}}\left( \theta  \right)\\
 = \frac{{{\lambda _0}}}{{NP}}{\sum\limits_{i = 1}^N {\sum\limits_{j = 1}^P {\left| {{G_\theta }\left( {{u_i}} \right)\left( {x,t} \right) - s\left( {x,t} \right)} \right|} } ^2} + \frac{{{\lambda _p}}}{{NQ}}\sum\limits_{i = 1}^N {\sum\limits_{j = 1}^Q {R_\theta ^i{{\left( {x,t} \right)}^2}} } 
\end{array}
\end{equation}
where ${\lambda}_o$ and ${\lambda}_p$ represent the ratio of the operator loss to the physical constraint loss in the overall loss, respectively. Based on the above, the complete pseudo-code for PI-DeepONet can be given as algorithm \ref{alg:pi_DeepONet}.
\begin{algorithm}
\caption{Physics-Informed DeepONet for Traffic State Estimation}
\label{alg:pi_DeepONet}
\textbf{Reuqire}:
\begin{itemize}
\setlength{\itemsep}{0pt}
    \item $\mathbf{s}_C \in \mathbb{R}^{M}$: Observed traffic state (speed) data at sparse spatiotemporal points.
    \item $\Omega \subset \mathbb{R}^2$: Full spatiotemporal domain.
    \item $\Theta = \{(x_j, t_j)\}_{j=1}^{P} \subset \Omega$: Collocation points for branch network input.
    \item ${U} = \{u_i\}_{i=1}^{N} \subset \mathcal{F}(\Omega)$: Set of random input functions.
    \item $v_f$: Free-flow speed (m/s), $\lambda_p \geq 0$: Physical loss weight, $\epsilon$: Numerical tolerance for convergence
\end{itemize}

\textbf{Ensure}:
\begin{itemize}
\setlength{\itemsep}{0pt}
    \item Trained operator $G_{\theta}: U(\Omega) \to \mathcal{S}(\Omega)$ mapping input functions to speed fields
\end{itemize} 
\begin{algorithmic}[0]
\State \textbf{Initialize} Branch network $g_{\theta_b}: \mathbb{R}^P \to \mathbb{R}^H$ and Trunk network $f_{\theta_t}: \mathbb{R}^2 \to \mathbb{R}^H$ with random weights $\theta = \{\theta_b, \theta_t\}$
\State \textbf{Generate} training dataset $\mathcal{D} = \{(\mathbf{U}_i, \mathbf{y}_k, s_C(\mathbf{y}_k))\}$ where:
    \[
    \mathbf{U}_i = [u_i(x_1, t_1), \dots, u_i(x_P, t_P)] \in \mathbb{R}^P \quad \forall u_i \in \mathcal{U},
    \]
    \[
    \mathbf{y}_k \in \Omega \text{ are observed points with labels } s_C(\mathbf{y}_k)
    \]
\State \textbf{Define} physical constraint via LWR-Greenshields model:
    \[
    \mathcal{L}_{phys}(v) = \frac{\partial v}{\partial t} + \left(2v - v_f\right)\frac{\partial v}{\partial x} = 0
    \]
\For{until $\left\| \nabla_{\theta} L_{total} \right\| < \epsilon$}
    \State \textbf{Sample} minibatch $\{(\mathbf{U}_i, \mathbf{y}_k, s_k)\}$ from $\mathcal{D}$
    \State \textbf{Forward Pass}:
        \[
        \mathbf{b}_i = g_{\theta_b}(\mathbf{U}_i), \quad \mathbf{t}_k = f_{\theta_t}(\mathbf{y}_k), \quad \hat{s}_k = \mathbf{b}_i \cdot \mathbf{t}_k
        \]
    \State \textbf{Data Loss}:
        \[
        L_{data} = \frac{1}{|\mathcal{B}|} \sum_{(\mathbf{U}_i, \mathbf{y}_k, s_k) \in \mathcal{B}} \left\| \hat{s}_k - s_k \right\|_2^2
        \]
    \State \textbf{Physical Loss} (evaluated at $Q$ randomly sampled interior points $\{\mathbf{z}_q\}_{q=1}^Q \subset \Omega$):
        \[
        \hat{v}_q = G_{\theta}(u_i)(\mathbf{z}_q), \quad R_q = \mathcal{L}_{phys}(\hat{v}_q),
        \]
        \[
        L_{phys} = \frac{1}{NQ} \sum_{i=1}^N \sum_{q=1}^Q \left\| R_q \right\|_2^2
        \]
    \State \textbf{Total Loss}:
        \[
        L_{total} = L_{data} + \lambda_p L_{phys}
        \]
    \State \textbf{Backward Pass}: Compute $\nabla_{\theta} L_{total}$ and update $\theta$ via Adam optimizer
\EndFor
\State \Return $G_{\theta}$
\end{algorithmic}
\end{algorithm}

\section{Experiment}\label{s4}
To evaluate the practical reliability of the proposed framework, we performed comparative experiments utilizing the widely recognized NGSIM dataset, as detailed in the subsequent sections. 

\subsection{Dataset}
\begin{figure}[pos=htbp]
    \centering
\includegraphics[width=0.9\textwidth]{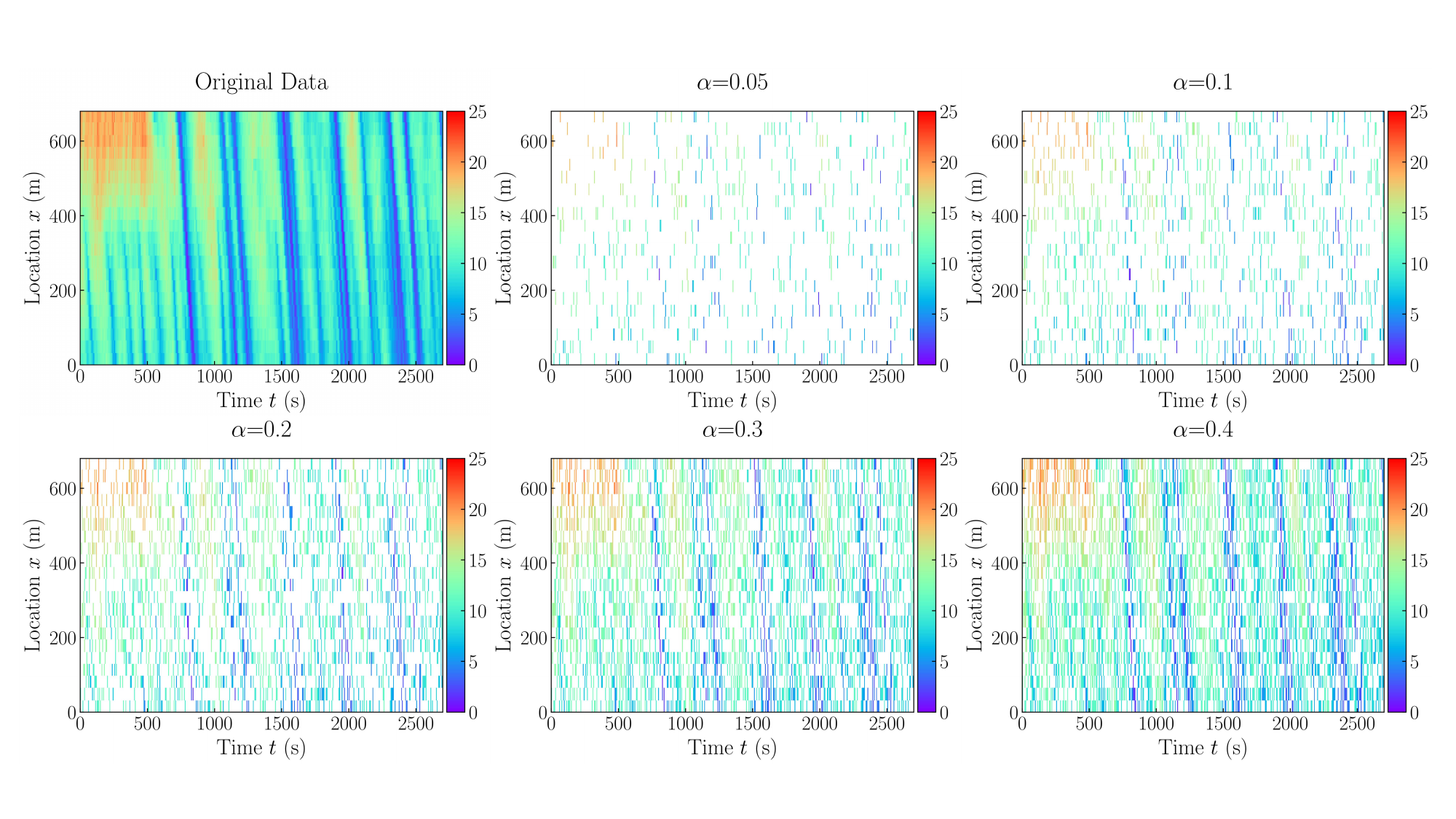}
\vspace{-5mm}
    \caption{\centering{Dataset: NGSIM highway}}
	\label{data}
\end{figure}

In this study, the NGSIM dataset is employed to assess the efficacy of the proposed NOs framework. This dataset encompasses positional and various status data for each vehicle within a surveillance area measuring 680 meters and spanning 2770 seconds, derived from camera video footage. The raw data were segmented into grid areas that account for vehicles across all lanes, employing a spatial resolution of 30 meters and a temporal interval of 1.5 seconds to derive flow data. The instantaneous speeds of vehicles were calculated to determine average speeds, while density was similarly computed. Following this methodology, a spatiotemporal grid of dimensions ${21 \times 1770}$ was ultimately constructed, as shown in Fig. \ref{data}. 

\subsection{Baseline and indicator}\label{setup}
In order to test the sophistication of the NOs model, the following comparison models were introduced in the study, mainly using a Python toolbox named PyPOTS for calling \footnote{https://github.com/WenjieDu/PyPOTS}:

NN: A neural network framework employed as a reference model for speed estimation through the learning of data characteristics \citep{lecun2015deep}.

BRITS: A bidirectional recurrent time series interpolation model designed to address estimation challenges in contexts where data is missing \citep{cao2018brits}.

GRUD: A sophisticated deep learning model that integrates global and recurrent units for the interpolation and estimation of missing values in time series data \citep{che2018recurrent}.

ETSFormer: An enhanced Transformer-based model utilized for speed estimation by leveraging spatiotemporal features \citep{woo2022etsformer}.

SAITS: A time series interpolation model that utilizes a self-attention mechanism to enhance estimation accuracy \citep{du2023saits}.

In our experiments, mean absolute error (MAE) and root mean square error (RMSE) are employed to evaluate the efficacy of the proposed framework, which was calculated using the following formulas: 
\begin{equation}
RMSE = \sqrt {\frac{1}{{{N_{val}}}}\sum\limits_{i = 1}^{{N_{val}}} {{{\left| {v\left( {{x_i},{t_i}} \right) - \widehat v\left( {{x_i},{t_i}} \right)} \right|}^2}} } \\
\end{equation}
\begin{equation}
MAE = \frac{1}{{{N_{val}}}}\sum\limits_{i = 1}^{{N_{val}}} {\left| {v\left( {{x_i},{t_i}} \right) - \widehat v\left( {{x_i},{t_i}} \right)} \right|} 
\end{equation}

\subsection{Experimental setup}\label{setup}
To ensure the accuracy of the physical data, this study calibrates the free stream velocity of the NGSIM US101 prior to conducting the experiments. The calibration process is supplemented by the selection of the root mean square error (RMSE) and the coefficient of determination (R²) as metrics for assessing the goodness of fit, yielding results of 4.154 and 0.721, respectively, as illustrated in Fig.\ref{calibration}. 

All models are executed on an Intel(R) Xeon(R) Platinum 8255C CPU operating at 2.50GHz and NVIDIA RTX 3090 GPUs, utilizing the PyTorch framework.Unless otherwise specified, the hyperparameters utilized in the experiment are shown in Table \ref{tab:parameters}. 

\begin{table}[pos=h]
  \centering
  \caption{Hyperparameters used in the experiment}
    \begin{tabular}{cccc}
    \toprule
    Type  & model & Hyperparameters & values \\
    \midrule
    \multirow{7}[2]{*}{Common} & \multirow{7}[2]{*}{\textbackslash{}} & Layers & 3 \\
          &       & \multicolumn{1}{c}{Hidden nodes} & 128 \\
          &       & \multicolumn{1}{c}{Learning rate} & 0.001 \\
          &       & \multicolumn{1}{c}{Activation function} & {GELU} \\
          &       & \multicolumn{1}{c}{Optimizer} & {Adma} \\
          &       & \multicolumn{1}{c}{Epoch} & 2000 \\
          &       & Criterion & MSE \\
    \midrule
    \multirow{6}[4]{*}{Particular} & \multirow{5}[2]{*}{DeepONet} & \multicolumn{1}{c}{Layers of Branch} & 3 \\
          &       & \multicolumn{1}{c}{Layers of Trunk} & 3 \\
          &       & Number of P & 100 \\
          &       & Number of u & 10 \\
          &       & Generator of u & GRF/Chebyshev \\
\cmidrule{2-4}          & PI-DeepONet  & $v_f$  & 19.965 m/s \\
    \bottomrule
    \end{tabular}%
  \label{tab:parameters}%
\end{table}%

\subsection{Results and discussion}
\subsubsection{Overall accuracy index}
\begin{figure}[pos=htbp]
    \centering
\includegraphics[width=0.8\textwidth]{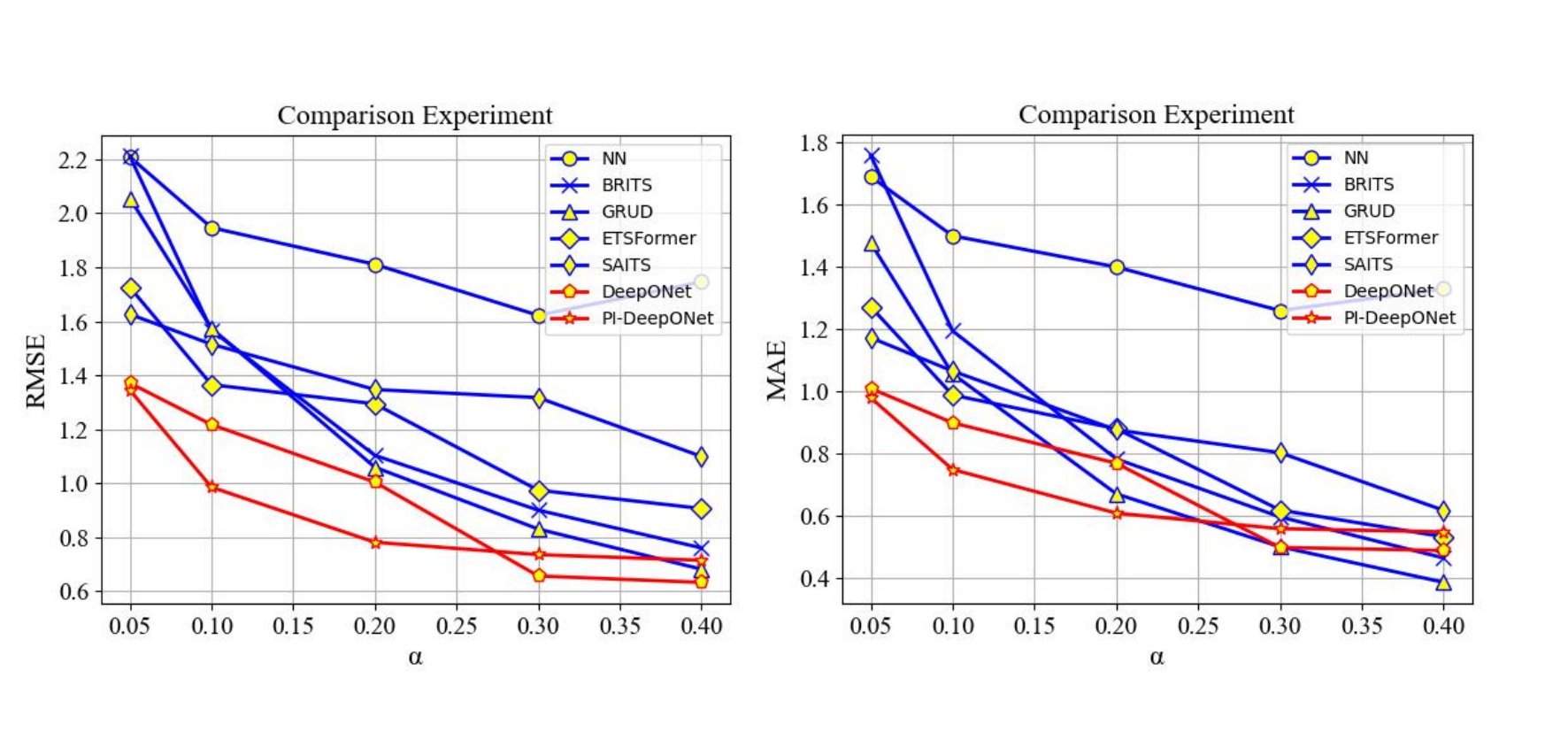}
    \caption{\centering{Speed estimation performance of NOs methods and baselines in NGSIM case.}}
	\label{performance}
\end{figure}

Table \ref{comparison} presents the evaluation results of the NOs model alongside five baseline models, focusing on prediction speed across various sampling rates. Fig.\ref{performance} provides a visual representation of the performance metrics assessed. The findings indicate that the performance of the NN is markedly inferior to that of the other models, attributable to their relatively simplistic architecture. In contrast, DeepONet and PI-DeepONet consistently outperform baseline in terms of RMSE at all sampling rates, with the most significant enhancement reaching 27.84\%. Given that RMSE is particularly sensitive to outlier values, the superior performance of NOs in this metric underscores its capability to effectively manage such mutated values. This proficiency is likely a result of the neural operators' approach of learning mappings between functions rather than relying solely on point-to-point mappings, which tend to yield smoother outputs, in contrast to the more variable nature of function-to-function mappings.

Furthermore, MAE of NOs is marginally lower than that of the baseline at the sampling rate of 40\% . This observation may be attributed to the ability to fit the function effectively of the baseline models within a specific dataset when it is sufficiently rich, although this often compromises the generalizability to other datasets of models. Returning to the internal analysis of NOs, the performance metrics reveal that DeepONet performs less effectively than PI-DeepONet at lower sampling rates; however, DeepONet begins to demonstrate superior performance at sampling rates greater than 30\%. This trend may be explained by the fact that the incorporation of physical equations at lower sampling rates provides additional information for the model to learn the operators, thereby compensating for data scarcity. Conversely, as the dataset becomes increasingly enriched, DeepONet is better able to approximate the true operators. In this context, the physical equations, which serve as approximation information, may inadvertently hinder the training process of model, resulting in DeepONet outperforming the PI-DeepONet under these conditions.

\begin{table}
  \centering
  \caption{Speed estimation comparison of NOs methods and baselines under randomly missing in NGSIM case}
  \label{comparison}
  \begin{tabular}{|c|c|c|c|c|c|c|c|c|c|c|}
    \hline
    \multirow{2}{*}{Model} & \multicolumn{2}{c|}{5\%} & \multicolumn{2}{c|}{10\%} & \multicolumn{2}{c|}{20\%} & \multicolumn{2}{c|}{30\%} & \multicolumn{2}{c|}{40\%} \\
    \cline{2-11}
     & RMSE & MAE & RMSE & MAE & RMSE & MAE & RMSE & MAE & RMSE & MAE \\
    \hline
    NN & 2.209 & 1.69 & 1.947 & 1.5 & 1.811 & 1.4 & 1.622 & 1.26 & 1.746 & 1.33 \\
    \hline
    BRITS & 2.212 & 1.758 & 1.567 & 1.195 & 1.103 & 0.784 & 0.9 & 0.598 & 0.76 & 0.466 \\
    \hline
    GRUD & 2.056 & 1.477 & 1.574 & 1.057 & 1.059 & 0.672 & 0.83 & 0.502 & 0.681 & \textbf{0.388} \\
    \hline
    ETSFormer & 1.726 & 1.272 & 1.365 & 0.989 & 1.294 & 0.881 & 0.974 & 0.62 & 0.906 & 0.534 \\
    \hline
    SAITS & 1.626 & 1.173 & 1.515 & 1.066 & 1.347 & 0.877 & 1.317 & 0.804 & 1.099 & 0.619 \\
    \hline \hline  
    DeepONet & 1.371 & 1.01 & 1.217 & 0.9 & 1.004 & 0.77 & \textbf{0.656} & \textbf{0.5} & \textbf{0.632} & 0.49 \\
    \hline
    Gain & 16.68\% & 13.90\% & 10.84\% &9.00\% & 5.19\% & -14.58\% & 20.96\% & 3.98\% & 7.20\% & -26.29\% \\
    \hline       
    PI-DeepONet & \textbf{1.345} & \textbf{0.98} & \textbf{0.985} & \textbf{0.75} & \textbf{0.781} & \textbf{0.61} & 0.735 & 0.56 & 0.714 & 0.55 \\
    \hline
    \hline
    Gain & 17.28\% & 16.45\% & 27.84\% & 24.17\% & 26.25\%  & 9.23\% & 11.44\% & -11.55\% & -4.85\% & -41.75\% \\
    \hline
  \end{tabular}
  \label{tab:ngsim_speed}
\end{table}

\subsubsection{Estimation results of global heatmap}

\begin{figure}[pos=htbp]
    \centering
\includegraphics[width=1\textwidth]{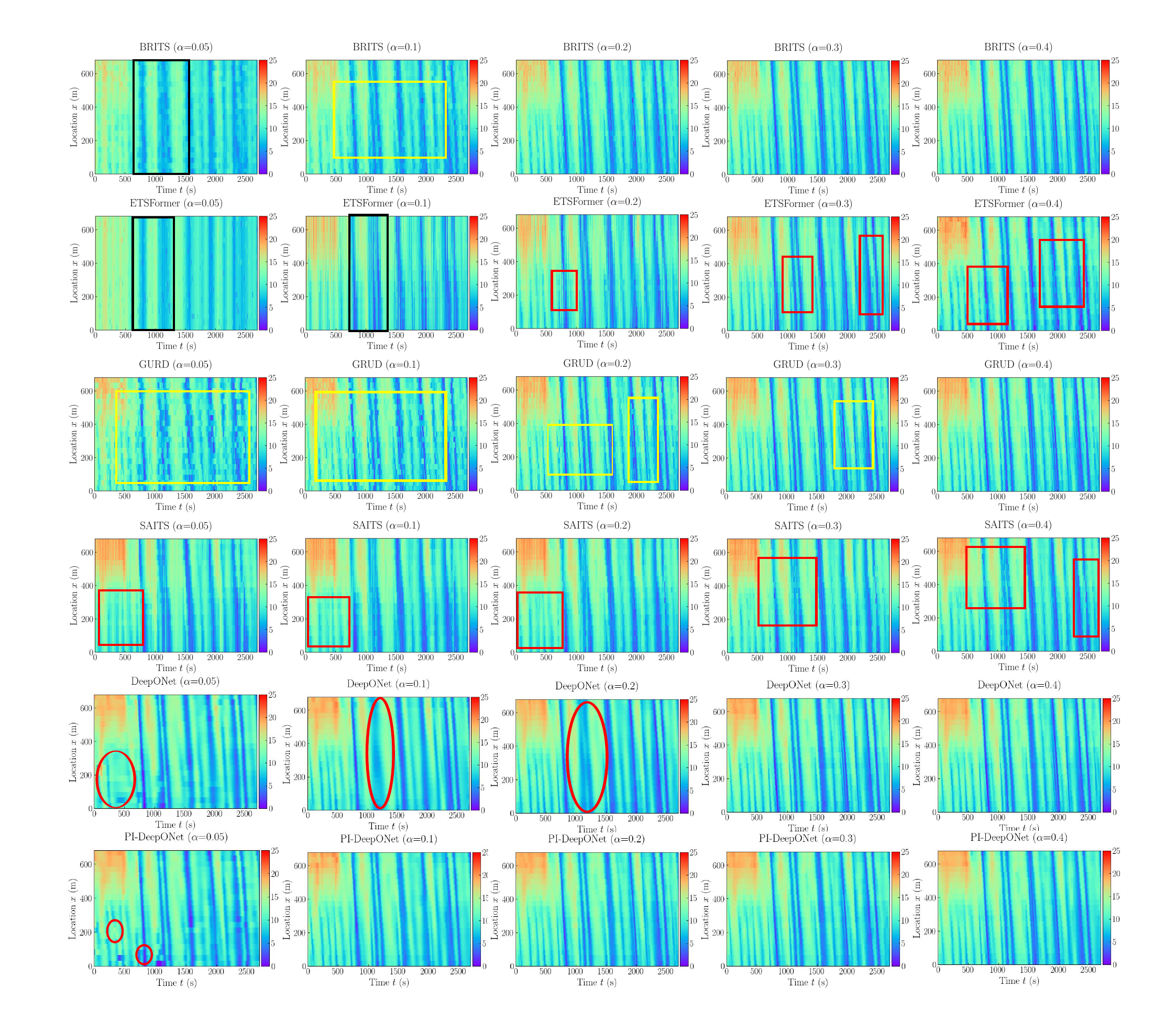}
    \caption{\centering{Speed estimation performance of NOs and baselines in NGSIM case.}}
	\label{prediction}
\end{figure}

Fig. \ref{prediction} presents a comparative heatmap analysis of NOs and  baselines at various sampling rates. The heatmaps indicate that at very low sampling rates ($\leq 0.1$), the estimations generated by NOs exhibit some discrepancies from the actual field data; however, the overall trend remains closely aligned. In contrast, baselines demonstrate notable deviations from expected outcomes. The following conclusions can be drawn from the analysis: 

1) At a sampling rate of 0.05, DeepONet struggles to effectively mitigate sparse wave phenomena due to insufficient data, as highlighted in the red elliptical box in the lower right corner. Nevertheless, it is capable of adequately addressing shock waves in accordance with fundamental physical principles. PI-DeepONet shows a comparative advantage in reducing sparse waves, attributed to the incorporation of a physical regularization term. However, it also exhibits a degree of smoothing in the context of shock waves, which may stem from the limitations of the linear assumptions inherent in the LWR-Greenshields model regarding the dynamics of shock wave evolution. Consequently, further training is necessary to accurately capture traffic wave patterns. At a sampling rate of 0.1, the aforementioned issues associated with DeepONet are mitigated to some extent, with PI-DeepONet nearly restoring all traffic wave characteristics present in the real data.

2) Although the baseline models are capable of partially reconstructing real field data at lower sampling rates, it generally exhibits a degree of physical inconsistency. For instance, the traffic waves predicted by BRITS and ESTFormer appear almost linear (as indicated by the black box), which contradicts the fundamental observation that traffic waves propagate backward over time. Similarly, while the predicted results of SAITS share a comparable overall shape with the actual result map, the sparse waves also exhibit near-linear characteristics (as shown in the red box in the lower right corner). Furthermore, the predictions from BRITS and GRUD display a significant tendency towards random fluctuations (as indicated by the yellow box).

As data continues to be enriched($\geq 0.2$), The predictions of the baselines still have some localized biases (red boxes), while NOs demonstrate improved capabilities in accurately reproducing the complete velocity field. Besides, the baselines exhibits a pronounced gridded characteristic, whereas NOs maintains better continuity of function-like while ensuring that the predictive outcomes encompass all relevant features. It is important to note that both models utilize the same gridded raw data and produce outputs in identical formats. This observation highlights the distinction between network learning functions and learning operators: NNs are designed to derive a function that characterizes the traffic state by directly fitting the labeled data, resulting in prediction outcomes that closely align with the individual gridded data points. In contrast, NOs approaches traffic state estimation from an infinite-dimensional function space, where the actual spatiotemporal function of speed serves as the supervisory label, and the gridded labeled data merely provide indirect observations. Consequently, the estimation results produced by the operator exhibit a greater overall continuity of the function.

\subsubsection{Estimation results of local evolution}
\begin{figure}[pos=htbp]
    \centering
\includegraphics[width=0.8\textwidth]{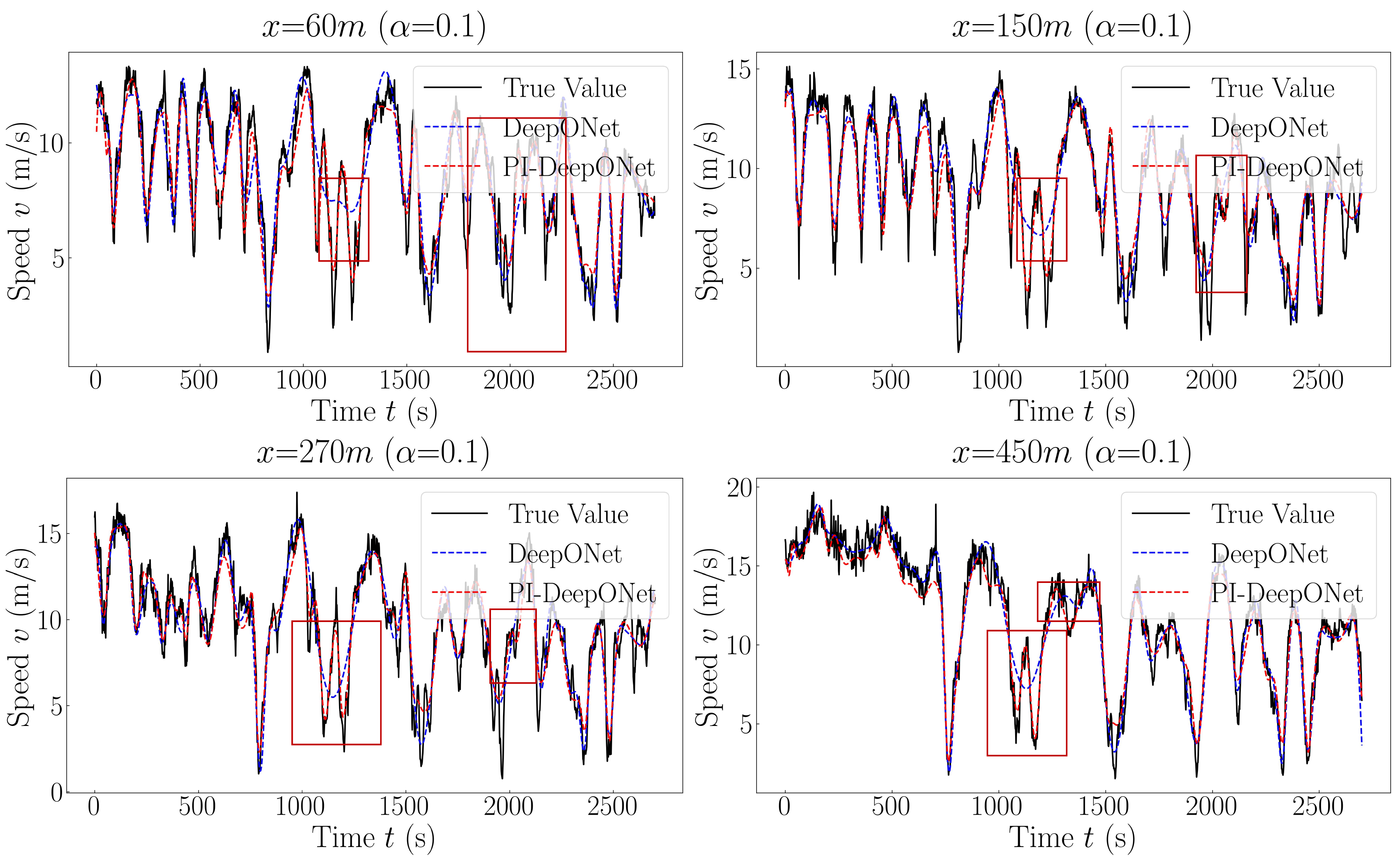}
    \caption{\centering{Estimation results of specific location.}}
	\label{location}
\end{figure}

\begin{figure}[pos=htbp]
    \centering
\includegraphics[width=0.8\textwidth]{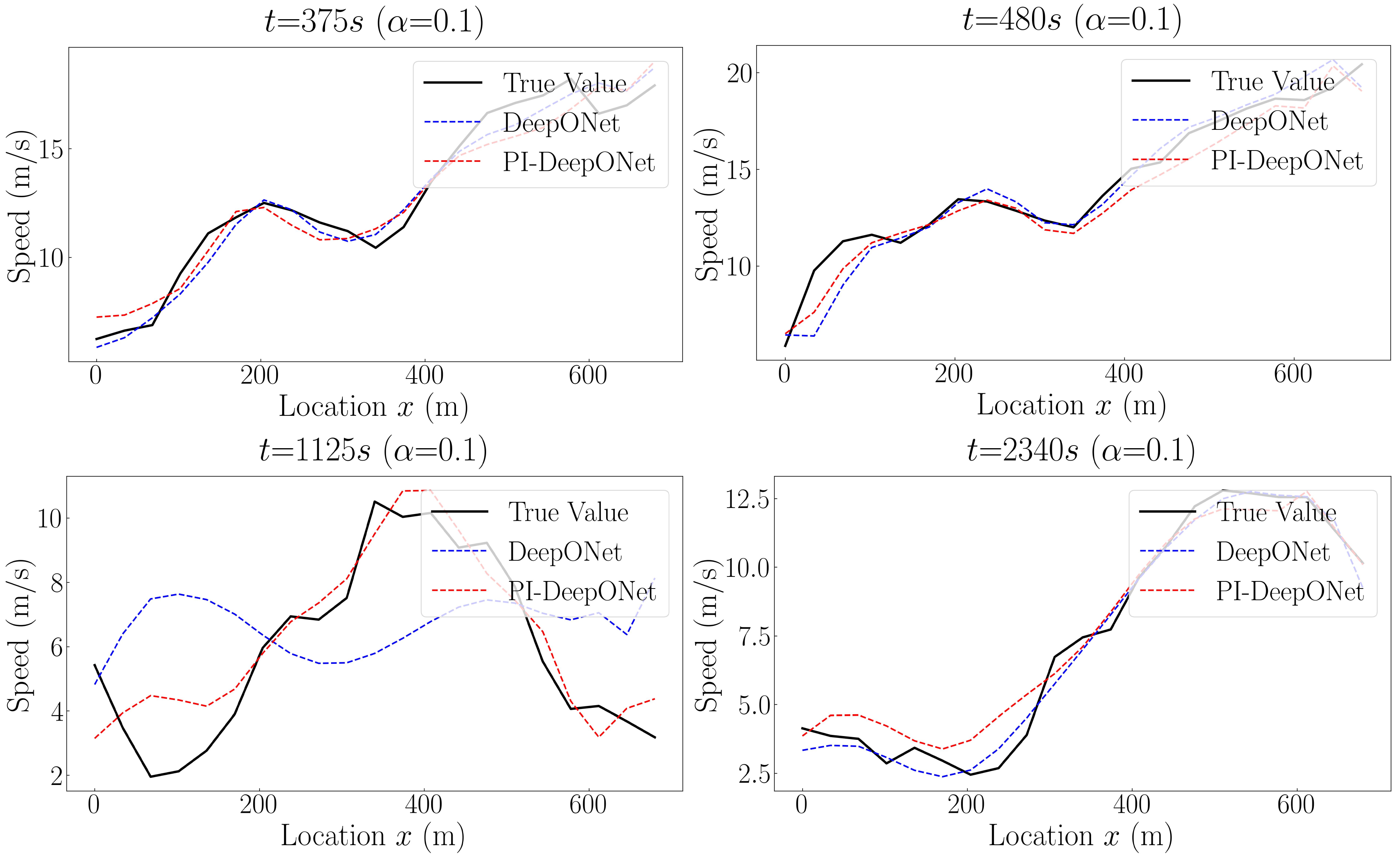}
    \caption{\centering{Estimation results of specific time.}}
	\label{time}
\end{figure}

Fig. \ref{location} and Fig.\ref{time} present a comparative analysis of DeepONet and PI-DeepONet at a designated location and time, utilizing a sampling rate of 10\%. In terms of positional profiles, both methodologies demonstrate superior predictive capabilities for velocity values; however, DeepONet exhibits some limitations in accurately forecasting velocity in certain localized change scenarios, as indicated by the red boxes in the figures. In contrast, the predictions of PI-DeepONet align closely with the actual values. This discrepancy in DeepONet can be attributed to its reliance on the universal approximation theorem to identify appropriate operators, resulting in an approximation error where the operators learned by the network deviate locally from the true operators. In contrast, PI-DeepONet incorporates physical information, thereby guiding the operator learning process and ensuring that the network adheres to fundamental traffic flow theories within the identified "operator clusters." The temporal profile exhibits characteristics analogous to the spatial profile; however, the gap in predictive accuracy is exacerbated due to the reduced dimensionality of the temporal profile. For example, at $t=1125s$ (see Fig.\ref{time}, bottom left), the prediction of DeepONet diverges significantly from the actual value, while the prediction of PI-DeepONet remains closely aligned with the true value. This further underscores the positive impact of integrating physical information on enhancing the generalization capabilities of the model.


\subsubsection{Sensitivity analysis}
\begin{figure}[pos=h]
    \centering
\includegraphics[width=1\textwidth]{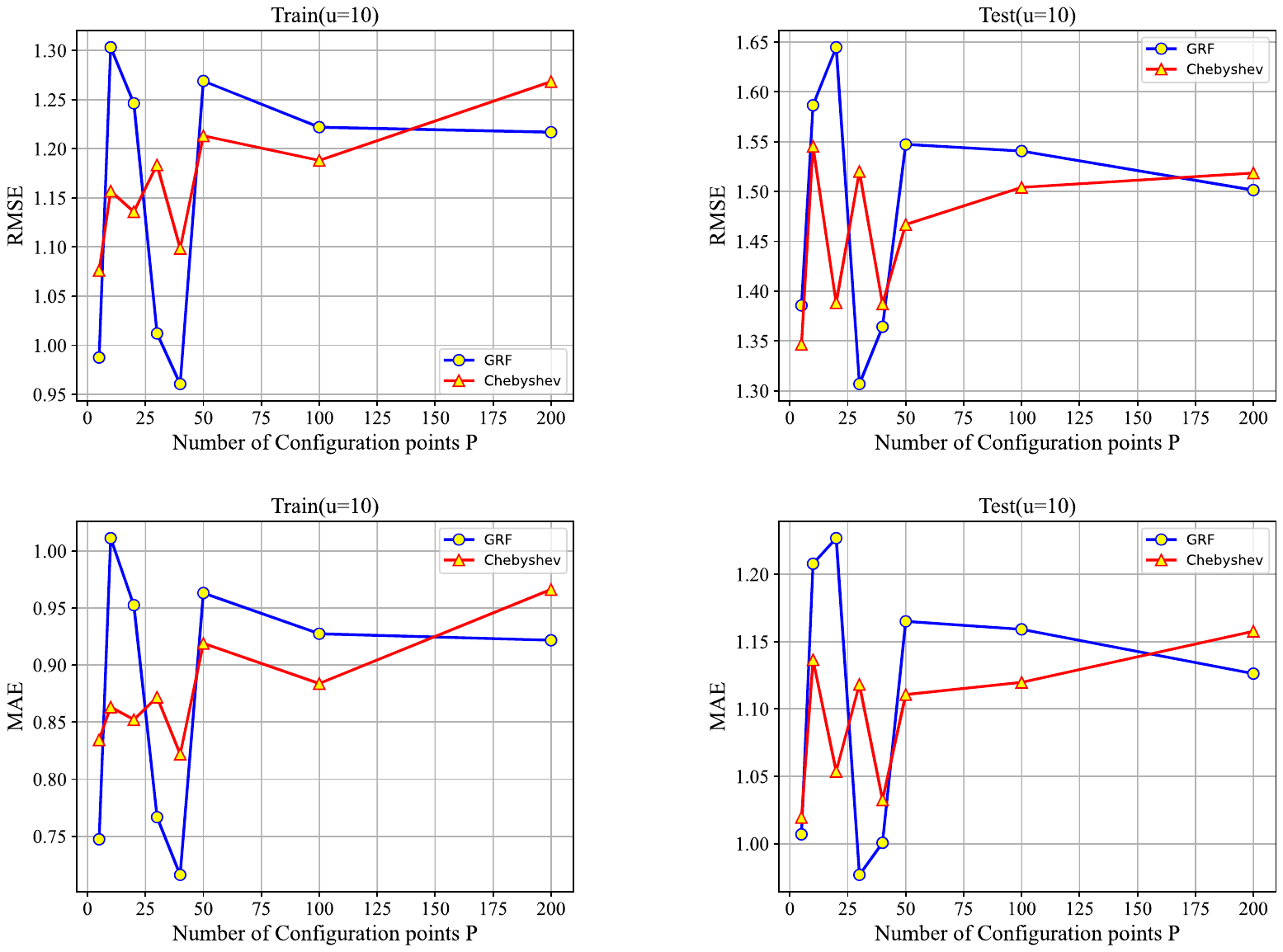}
    \caption{\centering{Error of NOs methods with different P.}}
	\label{curve of P}
\end{figure}

Fig.\ref{curve of P} presents the variations in error associated with two methods for generating input functions, analyzed across different quantities of sampling points, and detailed experimental data are presented in the Appendix Table\ref{tab:P_GRF} and \ref{tab:P_Chebyshev}. Overall, the influence of the number of configuration points $P$ on model error exhibits a nonlinear pattern characterized by an initial steep followed by a gradual leveling off, with some differences observed between the two generation methods. Specifically, when the number of P is relatively low $(P\le50)$, both methods demonstrate significant fluctuations in error. This phenomenon may be attributed to the insufficient number of P, which leads to a low sampling density of the function. Consequently, accurately reconstructing the characteristics of the input function within the generated training samples becomes challenging, resulting in pronounced fluctuations. Furthermore, the degree of fluctuation associated with the GRF method is generally greater than that of the Chebyshev method. This discrepancy may stem from the continuous smoothing assumption inherent in the GRF method, which necessitates a higher sampling density to effectively capture local features. As the number of P larger enough$(P>50)$, the errors for both methods tend to stabilize, indicating that the input function can be adequately characterized once the sampling density surpasses a certain threshold. Additionally, the error associated with the GRF method continues to exhibit a decreasing trend with increasing sampling density, suggesting that while the GRF method requires a greater number of configuration points, it ultimately facilitates a more precise characterization of the input function through denser sampling.

\begin{figure}[pos=h]
    \centering
\includegraphics[width=1\textwidth]{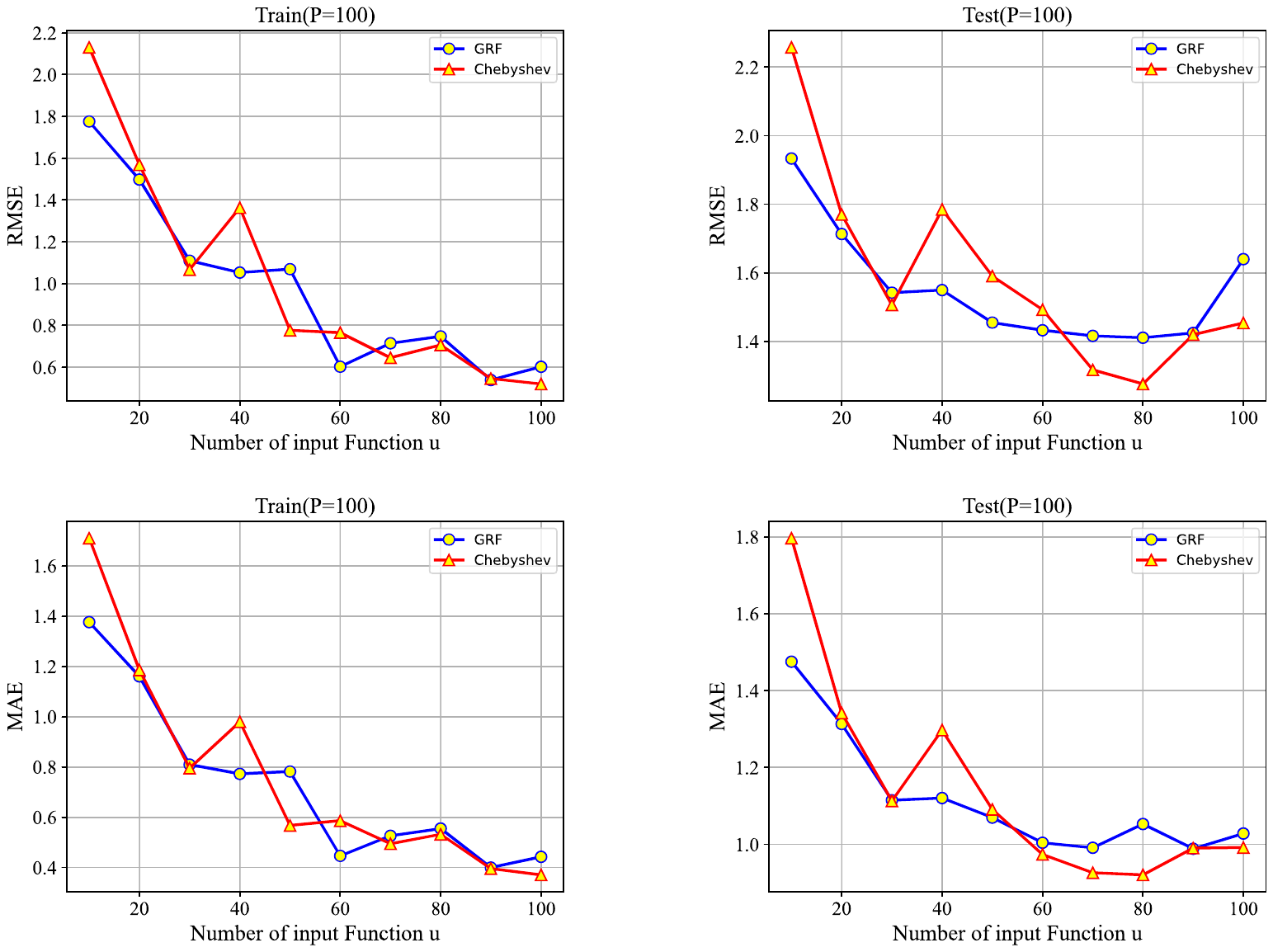}
    \caption{\centering{Error of NOs methods with different u.}}
	\label{curve of u}
\end{figure}

Fig.\ref{curve of u} presents the variations in error associated with two distinct methods for generating input functions u, across varying quantities of these functions, and detailed experimental data are presented in the Appendix Table\ref{tab:u_GRF} and \ref{tab:u_Chebyshev}. In contrast to the non-linear "steep and then smooth" trend observed in the configuration point experiments, both methods exhibit a pronounced decreasing trend in error as the number of input functions increases. However, some differences between the two generation methods persist. Specifically, when the number of input functions is relatively small $(\le30)$, the error associated with the Chebyshev method exceeds that of the GRF method; nevertheless, the Chebyshev method demonstrates a significantly faster rate of error reduction. This phenomenon may be attributed to the Chebyshev method's construction of the stochastic spatiotemporal function u through linear combinations of orthogonal basis functions. In scenarios where the number of input functions is limited, the generated input function set comprises a restricted number of basis function combinations, leading to underfitting in the model's ability to learn the superposition patterns of these functions. Increasing the number of input functions introduces a greater variety of basis function combinations, effectively alleviating this issue and facilitating a rapid decline in error. Despite the GRF method experiencing the same underfitting issue, its stochastic smoothing generation logic results in a lower degree of underfitting compared to the Chebyshev method. When the number of input functions is releatively lager $(>30)$, the error associated with the GRF method stabilizes, while the error for the Chebyshev method continues to exhibit a more pronounced decreasing trend, ultimately achieving a lower optimal error than that of the GRF method. Given that the operator $G$, as learned by NOs, in this study, represents an arbitrary spatiotemporal function mapping from $u(x,t)$ to $s(x,t)$, it follows that the trained model can effectively translate any spatiotemporal correlation u into the traffic state function $s(x,t)$ when utilized as input to the Branch network. From this perspective, the Chebyshev method may be deemed superior in generating a set of functions that possess a global representation within the operator $G$.

\section{Conclusion} \label{s5}
In this paper, we aim to address the complexities associated with sparse and noisy high-dimensional spatiotemporal data in TSE by introducing the Deep Operator Network (DeepONet) and the Physical Information Deep Operator Network (PI-DeepONet). The DeepONet framework reconceptualizes TSE as an operator learning challenge within a function space, aiming to learn the functional operator that relates the input function to the traffic state function. To enhance the model's ability to generalize across various input functions, this study employs GRF and Chebyshev polynomials to generate random functions. Subsequently, feature encoding within the function space is achieved through the utilization of branch networks and backbone networks, facilitating the mapping from sparse traffic state inputs to a comprehensive spatiotemporal state field. Moreover, the PI-DeepONet framework incorporates the conservation law of traffic flow and the Greenshields fundamental graph as physical regularization terms, thereby ensuring that the output traffic states adhere to the principles of traffic dynamics at the functional level. Empirical evaluations using real-world data demonstrate that both DeepONet and PI-DeepONet surpass baseline models across nearly all sampling rates, achieving maximum enhancements of 24.17\% and 27.84\% in MAE and RMSE, respectively. Notably, at lower sampling rates, the integration of physical constraints provides PI-DeepONet with a distinct advantage in terms of performance and physical consistency compared to DeepONet. Additionally, this study investigates the sensitivity variations associated with different input function generation methods. The findings reveal that GRF, owing to their smooth characteristics, yield superior performance at higher sampling densities (P $ \ge $ 100), while the linear approximation properties of Chebyshev polynomials result in reduced error when there is sufficient diversity of input functions (u $ \ge $ 60).

Future work will integrate multi-source data into the PI-DeepONet framework to enhance robustness and accuracy in capturing traffic dynamics. Additionally, developing real-time adaptive learning mechanisms will enable the model to dynamically update based on streaming data, improving responsiveness to changing traffic conditions.
\section*{Acknowledgement}
This research was supported by the project of the National Key R\&D Program of China (No. 2018YFB1601301), the National Natural Science Foundation of China (No. 71961137006, NO. 52302441).

\section*{Appendix}
\subsection*{Function Generation}
Algorithms \ref{alg:Generation of GRF} and \ref{alg:Generation of Chebyshev} show the generation of input functions via Gaussian random fields and Chebyshev polynomials, respectively.

\setcounter{algorithm}{0}
\begin{algorithm}[H]
\renewcommand{\thealgorithm}{A\arabic{algorithm}}
\caption{Gaussian Random Field-based Random Function Generation}
\label{alg:Generation of GRF}
\textbf{Require}:
\begin{itemize}
    \setlength{\itemsep}{0pt}
    \item $N$: Number of functions to generate 
    \item $L$: Length scale for filtering 
    \item $M$: Spatial dimension 
    \item $T$: Temporal dimension
\end{itemize}    
\textbf{Ensure}:
\begin{itemize}
    \setlength{\itemsep}{0pt}
    \item $U$: Set of generated functions $u(x,t)$
\end{itemize}    
\begin{algorithmic}[1]   
\State \textbf{Step 1: Compute filter parameters}
\State $\sigma_x \gets L \times M$  
\State $\sigma_t \gets L \times T$  

\State \textbf{Step 2: Generate functions}
\State $U \gets \emptyset$  
\For{$k \gets 0$ to $N-1$}
    \State $n \gets \text{normal}(0, 1, (M, T))$  
    \State $f \gets \text{GaussianFilter}(n, (\sigma_x, \sigma_t))$  
    \State $f \gets (f - \text{mean}(f))/\text{std}(f)$  
    \State $U \gets U \cup \{f\}$  
\EndFor

\State \Return $U$
\end{algorithmic}
\end{algorithm}

\setcounter{algorithm}{1}
\begin{algorithm}[H]
\renewcommand{\thealgorithm}{A\arabic{algorithm}}
\caption{Chebyshev Polynomial-based Random Function Generation}
\label{alg:Generation of Chebyshev}
\textbf{Require}:
\begin{itemize}
\setlength{\itemsep}{0pt}
    \item $N$: Number of functions to generate 
    \item $D$: Degree of Chebyshev polynomials 
    \item $M$: Spatial dimension 
    \item $T$: Temporal dimension
\end{itemize}
    
\textbf{Ensure}:
\begin{itemize}
\setlength{\itemsep}{0pt}
    \item $U$: Set of generated functions $u(x,t)$
\end{itemize}
\begin{algorithmic}[1]    
\State \textbf{Step 1: Generate normalized grid}
\State $x \gets \text{linspace}(-1, 1, M)$  
\State $t \gets \text{linspace}(-1, 1, T)$  

\State \textbf{Step 2: Precompute basis functions}
\State $B \gets \emptyset$  
\For{$i \gets 0$ to $D-1$}
    \For{$j \gets 0$ to $D-1$}
        \State $T_i \gets \text{Chebyshev}(i, x)$  
        \State $T_j \gets \text{Chebyshev}(j, t)$  
        \State $b \gets T_i \otimes T_j$  
        \State $B \gets B \cup \{b\}$
    \EndFor
\EndFor

\State \textbf{Step 3: Generate random functions}
\State $U \gets \emptyset$  
\For{$k \gets 0$ to $N-1$}
    \State $c \gets \text{randn}(|B|)$ 
    \State $f \gets \sum_{b \in B} c[b] \cdot b$ 
    \State $f \gets (f - \text{mean}(f))/\text{std}(f)$  
    \State $U \gets U \cup \{f\}$ 
\EndFor

\State \Return $U$
\end{algorithmic}
\end{algorithm}

\subsection*{Parameter calibration}
The calibration results of the Greenshields fundamental diagram model within PI-DeepONet is presented in Fig. \ref{calibration}. 
\setcounter{figure}{0}
\renewcommand{\thefigure}{A\arabic{figure}}
\begin{figure}
    \centering
\includegraphics[width=0.5\textwidth]{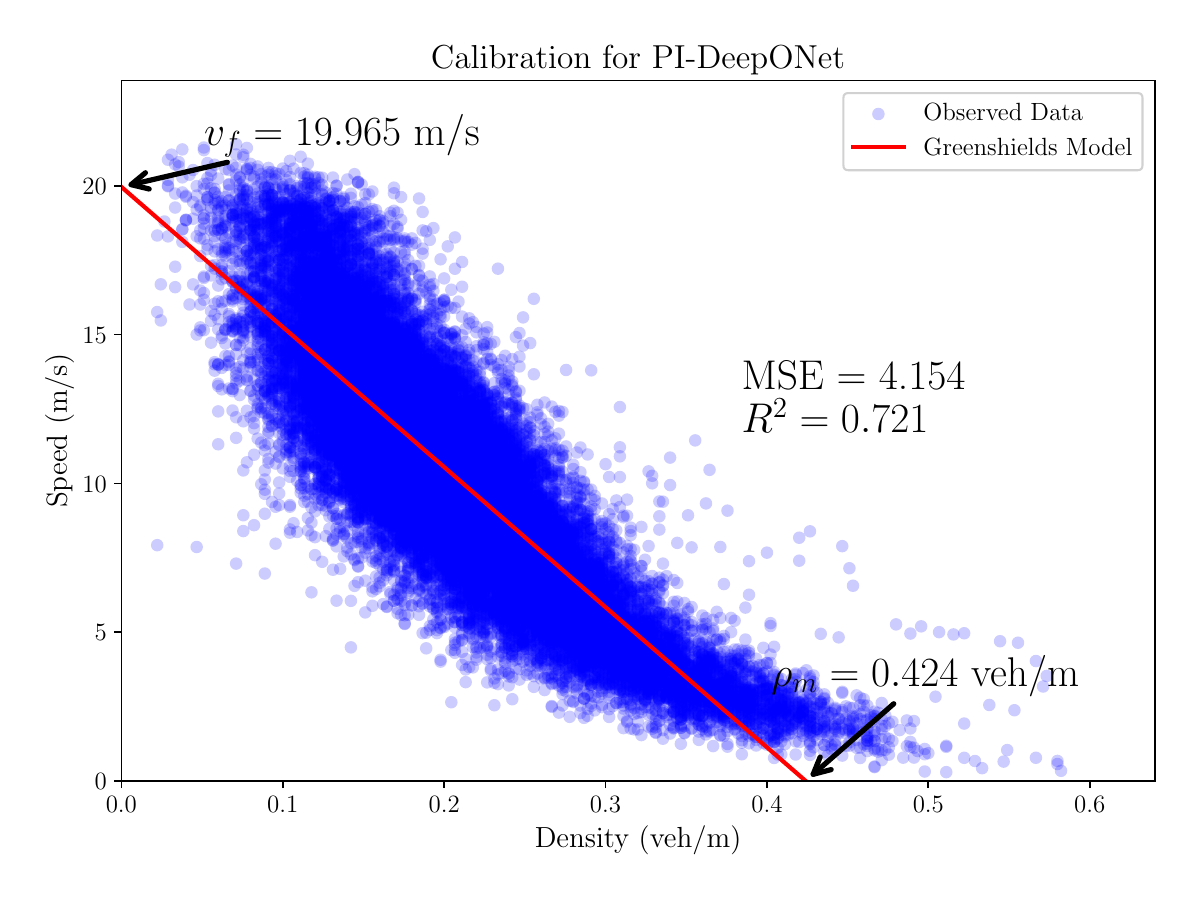}
    \caption{\centering{Calibration for PI-DeepONet.}}
	\label{calibration}
\end{figure}

\subsection*{Detailed experimental data for sensitivity analysis}
GRF and Chebyshev methods of input function generation are considered on the sensitivity experiments. Tables \ref{tab:P_GRF}–\ref{tab:P_Chebyshev} present the specific experimental data for both methods regarding the number of configuration points, corresponding to Fig.\ref{curve of P} .And  \ref{tab:u_GRF}–\ref{tab:u_Chebyshev} present the specific experimental data for both methods regarding the number of input functions, corresponding to Fig.\ref{curve of u}.

\setcounter{table}{0}
\renewcommand{\thetable}{A\arabic{table}}
\begin{table}[pos=h]
\belowrulesep=0pt
\aboverulesep=0pt
  \centering
  \caption{Specific experimental data in sensitivity experiments regarding the number of configuration points for the GRF method}
    \begin{tabular}{|c|c|c|c|c|c|c|c|c|}
    \toprule
    \multirow{2}[4]{*}{Number of P} & \multicolumn{2}{c|}{MSE} & \multicolumn{2}{c|}{RMSE} & \multicolumn{2}{c|}{MAE} & \multicolumn{2}{c|}{MAPE} \\
\cmidrule{2-9}          & Train & Test  & Train & Test  & Train & Test  & Train & Test \\
    \midrule
    5     & 0.9751 & 1.9198 & 0.9875 & 1.3856 & 0.7473 & 1.0069 & 10.94 & 15.01 \\
    \midrule
    10    & 1.698 & 2.5174 & 1.3031 & 1.5866 & 1.0112 & 1.2078 & 13.99 & 17.35  \\
    \midrule
    20    & 1.5528 & 2.7055 & 1.2461 & 1.6448 & 0.9525 & 1.2268 & 14.05 & 18.46 \\
    \midrule
    30    & 1.024 & 1.7074 & 1.0119 & 1.3067 & 0.7667 & 0.9768 & 11.78 & 14.97 \\
    \midrule
    40    & 0.9225 & 1.8609 & 0.9605 & 1.3642 & 0.7161 & 1.0006 & 10.75 & 15.34 \\
    \midrule
    50    & 1.6098 & 2.3942 & 1.2688 & 1.5473 & 0.9631 & 1.1650  & 13.91 & 17.29  \\
    \midrule
    100   & 1.4928 & 2.3733 & 1.2218 & 1.5406 & 0.9274 & 1.1591 & 13.86 & 17.48  \\
    \midrule
    200   & 1.4804 & 2.2541 & 1.2167 & 1.5014 & 0.9217 & 1.1262 & 13.3  & 16.72 \\
    \bottomrule
    \end{tabular}%
  \label{tab:P_GRF}%
\end{table}%

\setcounter{table}{1}
\renewcommand{\thetable}{A\arabic{table}}
\begin{table}[pos=h]
\belowrulesep=0pt
\aboverulesep=0pt
  \centering
  \caption{Specific experimental data in sensitivity experiments regarding the number of configuration points for the Chebyshev method}
    \begin{tabular}{|c|c|c|c|c|c|c|c|c|}
    \toprule
    \multirow{2}[4]{*}{Number of P} & \multicolumn{2}{c|}{MSE} & \multicolumn{2}{c|}{RMSE} & \multicolumn{2}{c|}{MAE} & \multicolumn{2}{c|}{MAPE} \\
\cmidrule{2-9}          & Train & Test  & Train & Test  & Train & Test  & Train & Test \\
    \midrule
    5     & 1.1579 & 1.8133 & 1.076 & 1.3466 & 0.8344 & 1.0194 & 11.99 & 15.07 \\
    \midrule
    10    & 1.3382 & 2.3881 & 1.1568 & 1.5454 & 0.8632 & 1.1365 & 13.11 & 17.64  \\
    \midrule
    20    & 1.29  & 1.9274 & 1.1358 & 1.3883 & 0.8521 & 1.0536 & 12.77 & 15.72 \\
    \midrule
    30    & 1.4008 & 2.3107 & 1.1835 & 1.5201 & 0.8718 & 1.1182 & 12.94 & 16.82 \\
    \midrule
    40    & 1.2064 & 1.9240  & 1.0984 & 1.3871 & 0.8219 & 1.0325 & 12.34 & 15.74 \\
    \midrule
    50    & 1.4717 & 2.1516 & 1.2131 & 1.4668 & 0.919 & 1.1107  & 12.77 & 15.94  \\
    \midrule
    100   & 1.4113 & 2.2623 & 1.188 & 1.5041 & 0.8838 & 1.1197 & 13.53 & 17.33  \\
    \midrule
    200   & 1.6083 & 2.3061 & 1.2682 & 1.5186 & 0.9662 & 1.1577 & 13.38 & 16.63 \\
    \bottomrule
    \end{tabular}%
  \label{tab:P_Chebyshev}%
\end{table}%

\setcounter{table}{2}
\renewcommand{\thetable}{A\arabic{table}}
\begin{table}[pos=h]
\belowrulesep=0pt
\aboverulesep=0pt
  \centering
  \caption{Specific experimental data in sensitivity experiments regarding the number of input function for the GRF method}
    \begin{tabular}{|c|c|c|c|c|c|c|c|c|}
    \toprule
    \multirow{2}[4]{*}{Number of u} & \multicolumn{2}{c|}{MSE} & \multicolumn{2}{c|}{RMSE} & \multicolumn{2}{c|}{MAE} & \multicolumn{2}{c|}{MAPE} \\
\cmidrule{2-9}          & Train & Test  & Train & Test  & Train & Test  & Train & Test \\
    \midrule
    10    & 3.1525 & 3.7382 & 1.7755 & 1.9334 & 1.3756 & 1.4751 & 20.32 & 22.84 \\
    \midrule
    20    & 2.2424 & 2.934 & 1.4975 & 1.7129 & 1.1596 & 1.3131 & 16.3  & 19.07 \\
    \midrule
    30    & 1.2305 & 2.3773 & 1.1093 & 1.5419 & 0.809 & 1.1147 & 11.84 & 16.56 \\
    \midrule
    40    & 1.1078 & 2.4004 & 1.0525 & 1.5493 & 0.7723 & 1.1205 & 11    & 16.22 \\
    \midrule
    50    & 1.1424 & 2.1155 & 1.0688 & 1.4545 & 0.7814 & 1.0696 & 11.37 & 15.73  \\
    \midrule
    60    & 0.3624 & 2.0512 & 0.602 & 1.4322 & 0.4462 & 1.0042 & 6.04  & 13.93 \\
    \midrule
    70    & 0.5104 & 2.0039 & 0.7144 & 1.4156 & 0.5259 & 0.9910  & 7.57  & 14.17 \\
    \midrule
    80    & 0.5583 & 2.0005 & 0.7472 & 1.4143 & 0.5548 & 1.0529 & 7.67  & 14.88 \\
    \midrule
    90    & 0.2899 & 2.0285 & 0.5384 & 1.4242 & 0.3996 & 0.9886 & 5.16  & 13.35 \\
    \midrule
    100   & 0.3623 & 2.6883 & 0.6019 & 1.6396 & 0.4425 & 1.0282 & 5.96  & 14.03 \\
    \bottomrule
    \end{tabular}%
  \label{tab:u_GRF}%
\end{table}%

\setcounter{table}{3}
\renewcommand{\thetable}{A\arabic{table}}
\begin{table}[pos=h]
\belowrulesep=0pt
\aboverulesep=0pt
  \centering
  \centering
  \caption{Specific experimental data in sensitivity experiments regarding the number of input function for the Chebyshev method}
    \begin{tabular}{|c|c|c|c|c|c|c|c|c|}
    \toprule
    \multirow{2}[4]{*}{Number of u} & \multicolumn{2}{c|}{MSE} & \multicolumn{2}{c|}{RMSE} & \multicolumn{2}{c|}{MAE} & \multicolumn{2}{c|}{MAPE} \\
\cmidrule{2-9}          & Train & Test  & Train & Test  & Train & Test  & Train & Test \\
    \midrule
    10    & 4.539 & 5.0979 & 2.1305 & 2.2579 & 1.7094 & 1.7964 & 25.78 & 27.78 \\
    \midrule
    20    & 2.4599 & 3.1326 & 1.5684 & 1.7699 & 1.1856 & 1.3417 & 17.73 & 20.18 \\
    \midrule
    30    & 1.1367 & 2.268 & 1.0662 & 1.506 & 0.7949 & 1.113 & 11.23 & 16.17 \\
    \midrule
    40    & 1.8578 & 3.1853 & 1.363 & 1.7847 & 0.9794 & 1.2972 & 14.13 & 18.75 \\
    \midrule
    50    & 0.6027 & 2.2584 & 0.7764 & 1.5901 & 0.5671 & 1.0908 & 8.42  & 15.93  \\
    \midrule
    60    & 0.5853 & 2.2269 & 0.7651 & 1.4923 & 0.5862 & 0.9739 & 7.71  & 13.58 \\
    \midrule
    70    & 0.4161 & 1.7338 & 0.645 & 1.3167 & 0.4945 & 0.9263  & 6.42  & 12.77 \\
    \midrule
    80    & 0.4988 & 1.6262 & 0.7063 & 1.2752 & 0.532 & 0.9206 & 7.4   & 13.08 \\
    \midrule
    90    & 0.2977 & 2.0145 & 0.5456 & 1.4193 & 0.3957 & 0.9903 & 5.31  & 13.43 \\
    \midrule
    100   & 0.2696 & 2.1125 & 0.5192 & 1.4534 & 0.3707 & 0.9918 & 5.27  & 13.57 \\
    \bottomrule
    \end{tabular}%
  \label{tab:u_Chebyshev}%
\end{table}%

\bibliographystyle{cas-model2-names}

\bibliography{main}

\end{sloppypar}
\end{document}